\documentclass[runningheads, a4paper]{llncs}
\usepackage{graphicx}
\usepackage{latexsym}
\usepackage{breakcites}
\usepackage{booktabs}
\usepackage{datetime}
\usepackage{subfig}
\usepackage{xspace}
\usepackage{fancyhdr}
\usepackage{amsmath}
\usepackage{xurl}
\usepackage{multirow}
\usepackage{adjustbox}
\usepackage{arydshln}
\usepackage{relsize}
\usepackage{wasysym}
\usepackage{multirow}
\usepackage{microtype}

\usepackage{cite}

\newcommand{\secref}[2][]{Section#1~\ref{sec:#2}}

\newcommand{\tabref}[2][]{Table#1~\ref{Tab:#2}}
\newcommand{\figref}[2][]{Figure#1~\ref{fig:#2}}

\begin{document}
\title{Rumour Detection via Zero-shot Cross-lingual Transfer Learning}
\author{Lin Tian\inst{1} \and
Xiuzhen Zhang\inst{1}\thanks{Corresponding author} \and
Jey Han Lau\inst{2}}
\authorrunning{L. Tian et al.}
%
\institute{RMIT University, Australia \\
\email{s3795533@student.rmit.edu.au, xiuzhen.zhang@rmit.edu.au}
\and
The University of Melbourne, Australia\\
\email{jeyhan.lau@gmail.com}}

\maketitle              
\begin{abstract}
Most rumour detection models for social media are designed 
for one specific language (mostly English). 
There are over 40 languages on Twitter and most languages lack annotated resources 
to build rumour detection models.
In this paper we propose a zero-shot cross-lingual transfer learning framework 
that can adapt a rumour detection model trained for a \textit{source 
language} to another \textit{target language}. Our framework utilises 
pretrained multilingual language models (e.g.\ multilingual BERT) and a 
self-training loop to iteratively bootstrap the creation of ``silver 
labels'' in the target language to adapt the model from the source 
language to the target language.
We evaluate our methodology on English and Chinese rumour datasets and 
demonstrate that our model substantially outperforms 
competitive benchmarks in \textit{both} {source and target language} 
rumour detection. 

\end{abstract}

\begin{table}[t]
\caption{An illustration of a COVID-19 rumour being circulated in 
  English, French and Italian on Twitter.}
  \label{Tab:sample_case}
\begin{center}
\scriptsize
\begin{adjustbox}{width=\textwidth}
\begin{tabular}{llc}
\hline
Date &
  Language &
  Tweet \\ \hline
04-02-2020 &
  English &
  \begin{tabular}{p{0.6\textwidth}}Bill Gates admits the vaccine will no doubt kill 700000 people.  The virus so far has killed circa 300000 globally. Can anyone explain to me why you would take a vaccine that kills more people than the virus it's desgined to cure?
  \end{tabular} \\ \hline
17-04-2020 &
  French &
  \begin{tabular}{p{0.6\textwidth}}et si bill gates etait le seul manipulateur de ce virus..  il veut moinsvde gens sur terre. veut vous vacciner et parle de pandemie depuis des années    c est quand meme fou cette citation, non? \#covid \#BillGates\end{tabular} \\ \hline
06-05-2020 &
  Italian &
  \begin{tabular}{p{0.6\textwidth}}Bill Gates:"la cosa più urgente nel mondo ora è il vaccino contro il Covid-19." I bambini africani che hanno ricevuto i vaccini di Bill Gates o sono morti o sono diventati epilettici. I vaccini di Bill Gates sono più pericolosi di qualsiasi coronavirus. \#BillGates \#Coronavirus\end{tabular} \\ \hline
\end{tabular}
\end{adjustbox}
\end{center}
\end{table}

\section{Introduction}

Online social media platforms provide an alternative means for the 
general public to access information. 
The ease of creating a social media account 
has the implication that rumours --- \textit{stories or statements 
with unverified truth value} \cite{allport1947psychology} --- can be 
fabricated by users and spread quickly on the platform.


To combat misinformation on social media, one may rely on fact checking 
websites such as \texttt{snopes.com} and \texttt{emergent.info}
to dispel popular rumours.  
Although manual evaluation is the most reliable way of identifying
rumours, it is time-consuming.

Automatic rumour detection is therefore desirable 
\cite{yang2012automatic,liu2015real}. 
Content-based methods focus on rumour detection using the textual 
content of messages and user comments.
Feature-based models exploit features other than text content,
such as author information and network propagation features, for rumour detection~\cite{ma2015detect,ma2017detect,mendoza2010twitter}. 
Most rumour detection models, however, are built for English 
\cite{shu2017fake,tian2020early}, and most 
annotated rumour datasets are also in English 
\cite{ma2017detect,kochkina2018all,shu2018fakenewsnet}.

Rumours can spread in different languages and across languages.  
Table~\ref{Tab:sample_case} shows an example  (untruthful) rumour about 
Bill Gates circulated on Twitter during the COVID-19 
pandemic.\footnote{\url{https://www.bbc.com/news/52847648}} The rumour 
is found not only in English but also in 
French and Italian.~\footnote{The following article clarifies several rumours 
surrounding Bill Gates: 
\url{https://www.reuters.com/article/uk-factcheck-gates-idUSKBN2613CK}.}

There are over 40 languages on 
Twitter\footnote{\url{https://semiocast.com/downloads/Semiocast_Half_of_messages_on_Twitter_are_not_in_English_20100224.pdf}} 
and most languages lack annotated resources for building rumour 
detection models.  Although we have seen recent successes  with deep 
learning based approaches for rumour 
detection~\cite{shu2017fake,ma2016detecting,zhou2019early,liu2018early,ma2015detect,bian2020rumor} 
most systems are monolingual and require annotated data to train a new model 
for a different language.

In this paper, we propose a zero-shot cross-lingual transfer 
learning framework for building a rumour detection system 
without requiring annotated data for a new language. 
Our system is cross-lingual in the sense that it can detect rumours in 
two languages based on one model.
Our framework first fine-tunes a multilingual pretrained language model 
(e.g.\ multilingual BERT) for rumour detection using annotated data for a source language (e.g. English), and 
then uses it to classify rumours on another target language (zero-shot 
prediction) to create ``silver'' rumour labels for the target language. 
We then use these silver 
labels to fine-tune the multilingual model further to adapt it to the 
target language. 



At its core, our framework is based on MultiFiT 
\cite{eisenschlos2019multifit} which uses a multilingual 
model (LASER~\cite{artetxe2019massively}) to perform zero-shot cross-lingual 
transfer from one language to another. An important difference is that 
we additionally introduce a self-training loop --- which iteratively 
refines the quality of the silver labels --- that can substantially 
improve rumour detection in the \textit{target language}.  Most 
interestingly, we also found that if we include the original gold labels 
in the source language in the self-training loop, detection performance 
in the \textit{source language} can also be improved, creating a rumour detection system that excels in \textit{both} source 
and target language detection.

To summarise, our contributions are: (1) we extend MultiFiT, a 
zero-shot cross-lingual transfer learning framework by introducing a 
self-training loop to build a cross-lingual model; and (2) we apply the 
proposed framework to the task of rumour detection, and 
found that our model substantially outperforms  
benchmark systems in both source and target language rumour detection.


\section{Related Work}


Rumour detection approaches can be divided into two major categories
according to the types of data used: text-based and non-text based.
Text-based methods focus on rumour detection using the textual content, 
which may include the original source document/message and user 
comments/replies.
A study~\cite{ma2016detecting} proposed a recursive neural network model to 
detect rumours.
Their model first clusters tweets by topics and then performs rumour 
detection at the topic level.  
Another study~\cite{shu2017fake} introduced linguistic 
features to represent writing styles and other features based on 
sensational headlines from Twitter and to detect misinformation.
To detect rumours as early as possible, a study~\cite{zhou2019early} incorporated 
reinforcement learning to dynamically decide how many responses are 
needed to classify a rumour. 
Some other study~\cite{tian2020early} explored the 
relationship between a source tweet and its comments by transferring 
stance prediction model to classify the veracity of a rumour.
Non-text-based methods utilise features such as user profiles or 
propagation patterns for rumour detection~\cite{ma2015detect,liu2018early}.
In this paper, we adopt the 
text-based approach to rumour detection.


Most studies on rumour detection focus on a specific 
social media platform or language (typically English).
Still there are a few exceptions that explore 
cross-domain/cross-lingual rumour detection or related tasks.
A study~\cite{wen2018cross} proposed a set of  10 hand-crafted cross-lingual and 
cross-platform features for rumour detection by capturing the similarity 
and agreement between online posts from different social media 
platforms.
Another study~\cite{mohtarami2019contrastive} introduced a contrastive learning-based 
model for cross-lingual stance detection using memory networks.
Different to these studies, we specifically focus on how to 
transfer learned knowledge from a source language 
to a target language for automatic rumour detection.

Transfer learning has been successfully applied to many natural language processing (NLP) tasks, 
where modern pretrained language models (e.g.\ BERT) are fine-tuned with annotated data for down-stream tasks  
~\cite{devlin2018bert,liu2019roberta,yin2020sentibert}.
Multilingual pretrained language models have also been explored. For example, BERT has a multilingual version trained using 104 languages of Wikipedia.~\footnote{\url{https://github.com/google-research/bert/blob/master/multilingual.md}.}  
A study~\cite{conneau2019unsupervised} incorporated RoBERTa's training 
procedure to pretrain a multilingual language model that produces  
sentence embeddings for 100 languages.
It is found that multilingual BERT is surprisingly good
at zero-shot cross-lingual transfer~\cite{pires2019multilingual}; in other words, it can be fine-tuned for a 
particular task in one language and used to make predictions in another 
language without any further training.
MultiFiT~\cite{eisenschlos2019multifit} was recently proposed, 
where a zero-shot  
cross-lingual transfer framework uses predicted labels from a 
fine-tuned multilingual model to train a monolingual model on the same 
task in a target language; the transfer learning objective is only to 
optimise the model for the target language. Different from MultiFiT, our objective
is to optimise models for both the target and source languages. 


Self-training~\cite{scudder1965probability} is an early semi-supervised 
learning approach that has been explored for a variety of NLP tasks, 
such as neural machine translation~\cite{he2019revisiting}, semantic 
segmentation~\cite{zou2018unsupervised}.
Self-training involves teacher and student models, where
the teacher model is trained with labelled data and then used to make 
predictions on unlabelled data to create more training data for training 
a student model.
The process is repeated for several iterations with the student model 
replacing the original teacher model at the end of each iteration, and 
through iterative refinement of the predicted labels the student model 
improves over time.
We apply self-training in a novel way to fine-tune pre-trained multilingual language models
for cross-lingual rumour detection, and show that the student model 
improves over time during the transfer.

\section{Methodology}
\label{sec:method}

We are interested in the task of rumour detection, and particularly how 
to do zero-shot cross-lingual transfer to build a cross-lingual rumour 
detection model. That is, we assume we have labelled rumours in one 
language (\textit{source}) where we can build a supervised rumour 
detection model, and the goal is to transfer the model to detect rumours 
in a second language (\textit{target}) without any labelled data in that 
second language. After transfer, it should have the ability to detect 
rumours in both languages (hence a multilingual model).
 We first describe the rumour classifier in \secref{rumour-detection}, 
and return to detail the cross-lingual transfer learning framework in 
\secref{cross-lingual}.

\subsection{Rumour Classifier}
\label{sec:rumour-detection}

\begin{figure}[t]
\centering
  \includegraphics[width=0.8\columnwidth]{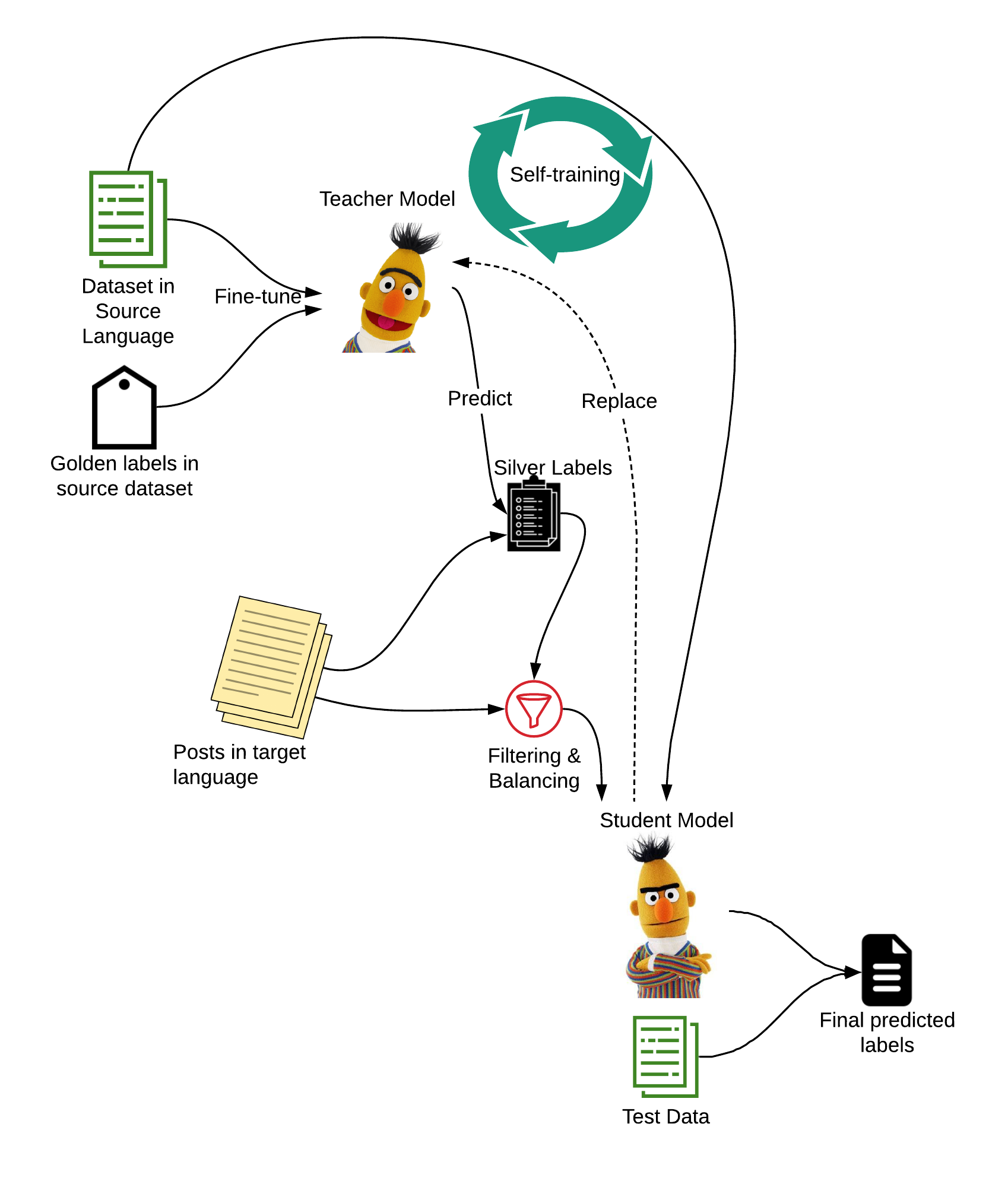}
  \caption{Proposed cross-lingual transfer framework.}
  \label{fig:model_archi}
\end{figure}

\begin{figure}[t]
\centering
  \includegraphics[width=0.8\columnwidth]{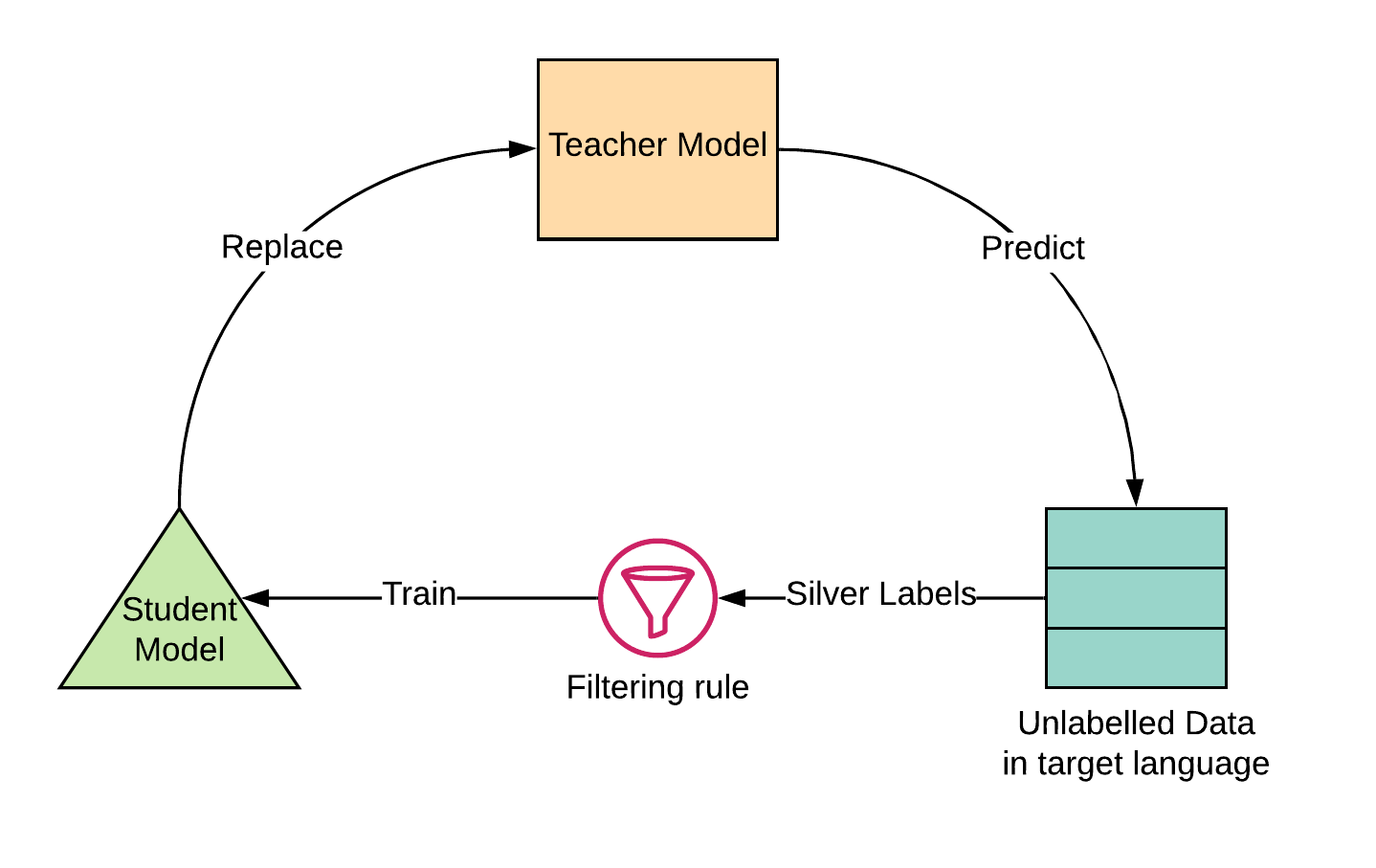}
  \caption{Self-training loop.}
  \label{fig:flow_chart}
\end{figure}

We focus on binary rumour detection, 
and follow previous studies to classify whether a microblog post constitutes a rumour or not based on crowd comments~\cite{zhou2019early,tian2020early,ma2016detecting}.



Given an initial post $s_i$ and its reactions $r_i$,\footnote{Reactions 
are replies and quotes. $r_i$ represents all reactions that can fit the 
maximum sequence length (384) for the pretrained model, concatenated 
together as a long string.} we feed them to a pretrained multilingual 
language model (we use multilingual BERT \cite{devlin2018bert} and 
XML-RoBERTa \cite{conneau2019unsupervised} in our experiments) 
as:\footnote{For XLM-RoBERTa, we have 
2 $[ SEP ]$ symbols between $s_i$ and $r_i$, following 
  \url{https://huggingface.co/transformers/model\_doc/xlmroberta.html\#transformers.XLMRobertaTokenizer.build\_inputs\_with\_special\_tokens}.}
\begin{equation*}
\left [ CLS \right ] + s_{i} + \left [ SEP \right ] + r_{i} + \left [ 
SEP \right ]
\end{equation*}
where $[CLS]$  and $[SEP]$ are special symbols used for classification 
and separating sequences \cite{devlin2018bert}.


We then take the contextual embedding of $[CLS]$ ($h_{[CLS]}$) and feed 
it to a fully-connected layer to perform binary classification of the 
rumour.
\begin{equation*}
  y_{i} = softmax\left ( W_{i}h_{\left [ CLS \right ]} + b_{s}\right )
\end{equation*}
Given ground truth rumour labels, the model is fine-tuned with standard 
binary cross-entropy loss. All parameters are updated except for the 
word embeddings (rationale detailed in the following section).


\subsection{Cross-lingual Transfer}
\label{sec:cross-lingual}

Our zero-shot cross-lingual transfer learning framework is based on 
MultiFiT~\cite{eisenschlos2019multifit}. MultiFiT works by first 
fine-tuning a multilingual model (e.g.\ LASER 
\cite{artetxe2019massively} is used in the original paper) for a task in 
a source language, and then applying it (zero-shot) to the same task in 
a target language to create silver labels. These silver labels are then 
used to fine-tune a monolingual model in the target language.  MultiFiT 
is shown to substantially improve document classification compared to 
zero-shot predictions by a series of multilingual models trained 
using only gold labels in the source language.\footnote{ 
Silver labels refer to the predicted labels in the target language, 
while gold labels refer to the real labels in the source language.}

We present our zero-shot cross-lingual transfer learning framework in 
\figref{model_archi}. One key addition that we make is a self-training 
loop that \textit{iteratively refines the quality of the adapted model}.  
In the original MultiFiT framework, the teacher model is a multilingual 
model, and the student model is a monolingual model in the target 
language.  As we are interested in multilingual rumour detection, the 
student model is a multilingual model in our case, although in our 
experiments (\secref{results}) we also present variations where the 
student model is a monolingual model.

Figure~\ref{fig:flow_chart} illustrates the self-training loop.
The student model is initialised using the teacher model (so both are 
multilingual models).  Once the student is trained, the teacher model in 
the next iteration will be replaced by the student model.

To reduce noise in the silver labels, we introduce 
a filtering and balancing procedure in the self-training loop. 
The procedure was originally introduced to image classification and shown to improve performance \cite{xie2020self}.
With the filtering procedure, 
instances with prediction confidence/probability 
lower than a threshold $p$ are filtered. 
The balancing procedure effectively drops some 
high confidence instances that pass the threshold to ensure that an equal number of positive (rumour) and negative (non-rumour) instances.

Following \cite{gururangan2020don}, we perform adaptive pretraining 
on the teacher model before fine-tuning it for the rumour detection
task.  That is,  we take the off-the-shelf pretrained multilingual model 
and further pretrain it using the masked language model objective on  
data in our rumour detection/social media domain.  In terms of 
pretraining data we use both the unlabelled rumour detection data 
(``task adaptive'') and externally crawled microblog posts (``domain 
adaptive'') in the target language.

The degree of overlap in terms of vocabulary between the source and target language varies depending on the language pair. If the overlap is 
low, after fine-tuning the source language subword embeddings would have 
shifted while the target language subword embeddings remained the same 
(due to no updates), creating a synchronisation problem between the 
subword embeddings and intermediate layers. We solve this issue by 
freezing the subword embeddings when we first fine-tune the teacher 
model; subsequent fine-tuning in the self-training loop, however, 
updates all parameters. We present ablation tests to demonstrate the 
importance of doing this in \secref{freeze}.

Aiming for a model that performs well for both the target and source languages,  we also introduce gold labels in the 
source language during self-training, i.e.\ we train the student using 
both the silver labels in the target language and the gold labels in the 
source language. This approach produces a well-balanced  
rumour detection model that performs well in both source and target 
languages, as we will see in \secref{results}.

\section{Experiments and Results}~\label{sec:exp}

We evaluated our cross-lingual transfer learning framework for rumour detection using three English and Chinese datasets. We formulate the problem as a binary classification task to distinguish rumours from non-rumours.

\subsection{Datasets}
Two English datasets Twitter15/16~\cite{ma2017detect} and PHEME~\cite{kochkina2018all}, and one Chinese dataset WEIBO~\cite{ma2016detecting} were used in our experiments. 
PHEME and WEIBO have two class labels, rumour and non-rumour.
For the Chinese WEIBO dataset, rumours are defined as ``a set of known rumours from the Sina community management center (http://service.account.weibo.com), which reports various misinformation"~\cite{ma2016detecting}. 
The original Twitter15/16 dataset~\cite{ma2017detect} has four classes, true rumour, false rumour, unverified rumour and non-rumour. 
We therefore extract tweets with labels ``false rumour" and ``non-rumour" from Twitter15/16 to match the definition of rumours and non-rumours of WEIBO and use the extracted data for experiments.
Table~\ref{Tab:Dataset} shows statistics of the experiment datasets.

To ensure fair comparison of the performance across all models,
for each dataset we reserved 20\% data as test and we split the rest in a ratio of 4:1 for training and validation partitions. The validation set was 
used for hyper-parameter tuning and early-stopping.
For the PHEME dataset, to be consistent with the experiment set up in the literature, we followed the 5-fold split from~\cite{ma2019detect}. 
For adaptive pretraining (\secref{cross-lingual}), we used an external 
set of microblogs data for English (1.6M posts; 
\cite{shu2018fakenewsnet}) and Chinese (39K 
posts).\footnote{\url{https://archive.ics.uci.edu/ml/datasets/microblogPCU}}

\begin{table}[t]
\caption{Rumour datasets.}
\label{Tab:Dataset}
\begin{center}
\begin{adjustbox}{width=0.6\linewidth}
\begin{tabular}{rrrr}
\toprule
&T15/16 & PHEME & WEIBO\\ \midrule
\#initial posts &1,154 &2,246 &4,664 \\
\#all posts &182,535 &29,387 &3,805,656 \\
\#users & 122,437 &20,529 &2,746,818 \\
\#rumours  & 575 &1,123&2,313 \\
\#non-rumours  &579 &1,123&2,351 \\
Avg. \# of reactions &279&26&247 \\
Max. \# of reactions &3,145&289&2,313 \\
Min. \# of reactions &74&12&10 \\ \bottomrule
\end{tabular}
\end{adjustbox}
\end{center}
\end{table}

\subsection{Experiment Setup}

We used multilingual BERT \cite{devlin2018bert} and XLM-RoBERTa 
\cite{conneau2019unsupervised} for the multilingual models, and implemented in PyTorch using the HuggingFace 
Libraries.\footnote{\url{https://github.com/huggingface}.}

For adaptive pretraining, we set batch size$=$8.
For the fine-tuning, we set batch size$=$16, maximum token length$=$384, 
and dropout rate$=$0.1. Training epochs vary between $3$--$5$ and 
learning rate in the range of \{1$e$-5, 2$e$-5, 5$e$-5\}; the best 
configuration is chosen based on the development data.
We also tuned the number of self-training iterations and $p$, the 
threshold for filtering silver labels (\secref{cross-lingual}), based on 
development.\footnote{For $p$ we search in the range of $0.94$--$0.96$.}
All experiments were conducted using 1$\times$V100 GPU. 

\begin{table}[t]
\caption{\label{Tab:result_table} Rumour detection results (Accuracy (\%)) for English to Chinese transfer. Each result 
is an average over 3 runs, and subscript denotes standard deviation.  monoBERT is a Chinese BERT model in this case.
Boldfont indicates optimal zero-shot performance.}
\begin{center}
\begin{tabular}{lc@{}c@{}c@{\;\;\;\;\;\;\;}c@{}c@{}c}
\toprule
\toprule
\multirow{2}{*}{\textbf{Model}} & \textbf{T15/16} &$\rightarrow$& \textbf{WEIBO} & \textbf{PHEME} &$\rightarrow$& \textbf{WEIBO} \\ 
                       & \textbf{Source}          && \textbf{Target}         & \textbf{Source}         && \textbf{Target}          \\ 
\midrule
\midrule
\multicolumn{5}{l}{Supervised}                          \\ 
\midrule
multiBERT$+$source &$95.8_{0.1}$  &&---  &$83.7_{0.5}$  &&---  \\ 
multiBERT$+$target &---  &&$93.9_{0.2}$  &--- &&$93.9_{0.2}$  \\ 
multiBERT$+$both &$94.8_{0.1}$ &&$93.0_{0.3}$ &$82.1_{0.8}$  &&$95.2_{0.2}$   \\ 
\hline
XLMR$+$source &$96.3_{0.4}$ &&---  &$82.8_{0.5}$  &&---  \\ 
XLMR$+$target &---  &&$94.8_{0.1}$  &---&&$94.8_{0.1}$  \\ 
XLMR$+$both &$95.5_{0.1}$  &&$92.2_{0.2}$  &$85.8_{1.9}$  &&$95.4_{0.1}$ \\ 
\midrule
\cite{ma2016detecting}$+$source &$83.5_{0.7}$ &&--- &$80.8_{0.4}$ &&---\\
\cite{liu2018early}$+$source &$85.4_{0.4}$ &&--- &$64.5_{1.0}$  &&---\\
\cite{tian2020early}$+$source &$87.2_{0.9}$ &&--- &$86.7_{1.5}$ && ---\\
\cite{bian2020rumor}$+$source &$96.3_{0.7}$ &&--- &--- &&--- \\
\midrule
\multicolumn{5}{l}{Zero-shot} \\ \midrule
multiBERT &---  &&$64.3_{2.1}$  &---  &&$65.9_{1.1}$ \\ 
XLMR &---  &&$64.7_{1.1}$  &---  &&$68.1_{1.0}$  \\ 
MF \cite{eisenschlos2019multifit} &---  &&$70.6_{0.4}$  &---  &&$61.1_{1.0}$  \\ 
\hline
MF-monoBERT &---  &&$67.5_{0.5}$  &---  &&$68.2_{0.4}$ \\ 
MF-monoBERT$+$ST &---  &&$\textbf{81.3}_{0.1}$  &---  &&$\textbf{79.0}_{0.2}$   \\ 
\hline
MF-multiBERT$+$ST &$61.3_{0.4}$  &&$78.6_{3.9}$  &$66.3_{1.5}$  &&$72.6_{1.9}$  \\ 
MF-multiBERT$+$ST$+$GL &$\textbf{96.6}_{0.2}$  &&$78.3_{0.8}$  &$83.0_{0.5}$  &&$74.3_{3.3}$   \\ 
\hline
MF-XLMR$+$ST  &$57.6_{1.0}$  &&$81.2_{0.1}$  &$62.1_{1.3}$  &&$77.4_{0.3}$   \\ 
MF-XLMR$+$ST$+$GL &$96.2_{0.1}$  &&$80.2_{0.2}$  &$\textbf{85.3}_{0.7}$  &&$77.2_{0.8}$  \\ 
\bottomrule
\bottomrule
\end{tabular}
\end{center}
\end{table}

\begin{table}[t]
\caption{\label{Tab:result_table2} Rumour detection results (Accuracy (\%)) for Chinese to English transfer. Each result 
is an average over 3 runs, and subscript denotes standard deviation.  
monoBERT is an English BERT model here.
}
\begin{center}
\begin{tabular}{lc@{}c@{}c@{\;\;\;\;\;\;\;}c@{}c@{}c}
\toprule
\toprule
\multirow{2}{*}{\textbf{Model}} & \textbf{WEIBO} &$\rightarrow$& \textbf{T15/16} & \textbf{WEIBO} &$\rightarrow$& \textbf{PHEME} \\ 
                       & \textbf{Source}          && \textbf{Target}         & \textbf{Source}         && \textbf{Target}  \\ 
\midrule
\midrule
\multicolumn{5}{l}{Supervised}                          \\ 
\midrule
multiBERT$+$source &$93.9_{0.1}$  &&---  &$93.9_{0.2}$  &&---  \\ 
multiBERT$+$target &---  &&$95.8_{0.1}$  &---  &&$83.7_{0.5}$  \\ 
multiBERT$+$both &$93.0_{0.3}$  &&$94.8_{0.1}$  &$95.2_{0.2}$  &&$82.1_{0.8}$  \\ 
\hline
XLMR$+$source &$94.8_{0.1}$  &&---  &$94.8_{0.1}$ &&---  \\ 
XLMR$+$target &---  &&$96.3_{0.4}$  &---  &&$82.8_{0.5}$  \\ 
XLMR$+$both &$92.2_{0.2}$  &&$95.5_{0.1}$  &$95.4_{0.1}$  &&$85.8_{1.9}$  \\ 
\midrule
\cite{ma2016detecting}$+$source &$91.0_{0.1}$ &&--- &$91.0_{0.1}$ &&--- \\
\cite{liu2018early}$+$source &$92.1_{0.2}$ &&--- &$92.1_{0.2}$ &&--- \\
\cite{bian2020rumor}$+$source &$96.1_{0.4}$ &&--- &$96.1_{0.4}$ &&--- \\
\midrule
\multicolumn{5}{l}{Zero-shot} \\ \midrule
multiBERT &---  &&$60.8_{1.3}$  &---  &&$67.2_{0.4}$  \\ 
XLMR &---  &&$73.9_{0.8}$  &---  &&$69.0_{1.5}$  \\ 
MF \cite{eisenschlos2019multifit} &---  &&$73.4_{1.4}$  &---  &&$64.1_{2.0}$  \\ 
\hline
MF-monoBERT &---  &&$64.7_{0.1}$  &---  &&$70.7_{0.7}$  \\ 
MF-monoBERT$+$ST &---  &&$\textbf{85.7}_{0.4}$  &---  &&$\textbf{78.9}_{0.6}$  \\ 
\hline
MF-multiBERT$+$ST &$55.1_{1.8}$  &&$82.2_{1.2}$  &$66.0_{1.0}$  &&$72.6_{1.5}$  \\ 
MF-multiBERT$+$ST$+$GL  &$97.0_{0.1}$  &&$80.9_{1.5}$  &$95.8_{0.5}$  &&$73.4_{0.4}$  \\ 
\hline
MF-XLMR$+$ST  &$52.4_{0.3}$  &&$83.0_{1.0}$  &$62.7_{0.5}$  &&$75.4_{0.4}$  \\ 
MF-XLMR$+$ST$+$GL &$\textbf{97.6}_{0.1}$  &&$81.3_{0.1}$  &$\textbf{95.9}_{0.5}$  &&$77.9_{1.1}$  \\ 
\bottomrule
\bottomrule
\end{tabular}
\end{center}
\end{table}

\begin{table}[t]
\caption{\label{Tab:f1_results} Rumour detection results (F$_1$ score (\%)) for both the source and target languages. ``R" and ``NR'' denote the rumour and non-rumour classes respectively. }
\begin{center}
\begin{adjustbox}{max width=\linewidth}
\begin{tabular}{l@{\;\;\;\;\;\;\;}c@{}c@{}c@{\;\;\;\;\;\;\;}c@{}c@{}c@{\;\;\;\;\;\;\;}c@{}c@{}c@{\;\;\;\;\;\;\;}c@{}c@{}c}
\toprule
\toprule
\multirow{2}{*}{\textbf{}} & \textbf{T15/16} &$\rightarrow$& \textbf{WEIBO} & \textbf{PHEME} &$\rightarrow$& \textbf{WEIBO} & \textbf{WEIBO} &$\rightarrow$& \textbf{T15/16} & \textbf{WEIBO} &$\rightarrow$& \textbf{PHEME} \\ 
                       & \textbf{Source}          && \textbf{Target}         & \textbf{Source}         && \textbf{Target}        & \textbf{Source}         && \textbf{Target}         & \textbf{Source}       && \textbf{Target}        \\ 
\midrule
\midrule
\multicolumn{9}{l}{MF-multiBERT+ST+GL}                          \\ 
\midrule
\textbf{R} &$96.6$  &&$79.1$  &$84.2$  &&$75.5$  &$96.9$  &&$83.3$  &$94.3$  &&$75.2$  \\ 
\textbf{NR} &$96.1$  &&$77.2$  &$79.5$ &&$73.1$  &$97.0$  &&$76.8$  &$94.4$  &&$70.3$  \\ 
\midrule
\multicolumn{9}{l}{MF-XLMR+ST+GL} \\ \midrule
\textbf{R} &$96.6$  &&$83.1$  &$85.2$  &&$81.6$  &$97.4$  &&$82.9$  &$96.9$  &&$74.4$  \\ 
\textbf{NR} &$95.6$  &&$76.1$  &$85.7$ &&$74.5$  &$96.8$  &&$81.1$  &$95.0$  &&$81.4$  \\ 
\bottomrule
\bottomrule
\end{tabular}
\end{adjustbox}
\end{center}
\end{table}

\subsection{Results}
\label{sec:results}

For our results, we show cross-lingual transfer performance from English 
to Chinese and vice versa.  As we have two English datasets (T15/16 and 
PHEME) and one Chinese dataset (WEIBO), we have 
four sets of results in total: T15/16$\rightarrow$WEIBO, 
  PHEME$\rightarrow$WEIBO, WEIBO$\rightarrow$T15/16 and 
WEIBO$\rightarrow$PHEME.
We evaluate rumour detection performance using accuracy, and present the 
English$\rightarrow$Chinese results in 
Table~\ref{Tab:result_table} and Chinese$\rightarrow$English in Table~\ref{Tab:result_table2} respectively. 
All performance is 
an average over 
3 runs with different random seeds. 

We include both supervised and zero-shot baselines in our 
experiments.  For the supervised benchmarks, we trained multilingual BERT 
and XLM-RoBERTa using: (1) source labels; (2) target labels; and (3) 
both source and target labels.
The next set of supervised models are state-of-the-art monolingual  
rumour detection models: (1) \cite{ma2016detecting} is a neural model 
that processes the initial post and crowd comments with a 2-layer gated 
recurrent units; (2) \cite{liu2018early} uses recurrent and 
convolutional networks to model user metadata (e.g.\ followers count) in 
the crowd responses;\footnote{Following the original paper, only a 
maximum of 100 users are included.} (3) \cite{tian2020early} uses BERT 
to encode comments (like our model) but it is pre-trained with stance 
annotations; and (4) \cite{bian2020rumor} uses bidirectional graph 
convolutional networks to model crowd responses in the propagation path.  
Note that we only have English results (T15/16 and PHEME) for 
\cite{tian2020early} as it uses stance annotations from SemEval-2016 
\cite{mohammad2016semeval}, and only T15/16 and WEIBO results for 
\cite{bian2020rumor} as PHEME does not have the propagation network 
structure. To get user metadata for \cite{liu2018early}, we crawled user 
profiles via the Twitter 
API.\footnote{\url{https://developer.twitter.com/en/docs/twitter-api/v1}}

For the zero-shot baselines, multilingual BERT and 
XLM-RoBERTa were trained using the source labels and applied to the target 
language (zero-shot predictions); subword embeddings are frozen during 
fine-tuning for these zero-shot models.
We also include the original MultiFiT model 
\cite{eisenschlos2019multifit}, which uses 
LASER~\cite{artetxe2019massively} as the multilingual model (teacher) 
and a pretrained quasi-recurrent neural network language model  
\cite{bradbury2016quasi} as the monolingual model (student).\footnote{The 
monolingual student model is pretrained using Wikipedia in the target 
language.}


We first look at the supervising results. XLM-RoBERTa (``XLMR'') is 
generally better (marginally) than multilingual BERT (``multiBERT'').  
In comparison, for the monolingual rumour 
detection models, \cite{bian2020rumor} has the best performance overall 
(which uses network structure in addition to crowd comments), although 
XLM-RoBERTa and multilingual BERT are not far behind.

Next we look at the zero-shot results. Here we first focus on target 
performance and baseline models.
The zero-shot models (``multiBERT'' and  XLMR'') 
outperform the MultiFiT baseline (``MF'') in 2--3 out of 
4 cases,  challenging the original findings in 
  \cite{eisenschlos2019multifit}.  When we replace the teacher model 
with multilingual BERT and the student model with monolingual BERT 
(``MF-monoBERT''), we found mixed results compared to MultiFiT (``MF''): 
2 cases improve but the other 2 worsen. When we incorporate the 
  self-training loop (``MF-monoBERT$+$ST''), however, we see marked 
improvement in all cases --- the largest improvement is seen in 
WEIBO$\rightarrow$T15/16 (Chinese to English, Table~\ref{Tab:result_table2}), from 64.7\% to 85.7\% --- demonstrating the 
benefits of iteratively refining the transferred model.  These results 
set a new state-of-the-art for zero-shot cross-lingual transfer 
learning for our English and Chinese rumour detection datasets. That 
said, there is still a significant gap (10$+$ accuracy points) compared 
to supervised models, but as we see in \secref{few-shot} the gap 
diminishes quickly as we introduce some ground truth labels in the 
target domain.

We now discuss the results when we use a multilingual model for the student model, i.e.\ replacing it with either multilingual BERT (``MF-multiBERT$+$ST'') or XLM-RoBERTa 
(``MF-XLMR$+$ST''), which turns it into a \textit{multilingual} rumour detection system (i.e.\ after fine-tuned it can detect rumours in both source and target language). Similar to the supervised results, we see that the latter (``MF-XLMR$+$ST'') is a generally better 
multilingual model. Comparing our best 
multilingual student model (``MF-XLMR$+$ST'') to the monolingual student model 
(``MF-monoBERT$+$ST'') we see only a small drop in the target performance (about 
1--4 accuracy points depending on domain), demonstrating that the multilingual  rumour detection system is competitive to the monolingual detection system in the target language.

For the source performance, we see a substantial drop (20--40 accuracy 
points) after cross-lingual transfer (e.g.\ ``XLMR$+$source'' vs.  
``MF-XLMR$+$ST''), implying there is catastrophic forgetting
~\cite{french1999catastrophic,wiese2017neural,kirkpatrick2017overcoming,xu2020forget} --- the 
phenomenon where adapted neural models ``forget'' and perform poorly in 
the original domain/task. When we incorporate gold labels in the source 
domain in the self-training loop (``MF-multiBERT$+$ST$+$GL'' or 
``MF-XLMR$+$ST$+$GL''), we found a surprising observation: not only was 
catastrophic forgetting overcame, but the source performance actually 
surpasses some supervised monolingual models,  e.g.\  
``MF-multiBERT$+$ST$+$GL'' and ``MF-XLMR$+$ST$+$GL'' outperform 
\cite{bian2020rumor} in T15/16 (96.6\% vs.\ 96.3\%) and WEIBO (97.6\% vs.\ 
96.1\%) respectively, creating a new state-of-the-art for rumour detection 
   in these two domains. One explanation is that the transfer learning 
framework maybe functioning like a unique data augmentation technique 
that creates additional data in a different language (unique in the 
sense it works only for improving multilingual models).  Note that 
incorporating the gold labels generally does not hurt the \textit{target 
performance} -- e.g.\ comparing ``MF-XLMR$+$ST'' with 
``MF-XLMR$+$ST$+$GL'' we see a marginal dip in 2 cases, but in 2 other 
cases we see similar or improved performance --- which 
shows that this is an effective approach for building multilingual models.

We further examine class-specific performance of our best models.
The F$_1$ scores of rumour and non-rumour classes are presented in Table~\ref{Tab:f1_results}.
For this binary classification task with relatively balanced class distributions, not surprisingly we observe that our models have reasonably good performance in both the rumour and non-rumour classes;
lowest F$_1$ score is 70.3\% of MF-multiBERT+ST+GL (Chinese to English transfer ) for non-rumours in PHEME. 
That said, performance of the rumour class is generally better than that of the non-rumour class in both the source and target languages (the only exception is Chinese to English transfer on PHEME). 

\subsection{Adaptive Pretraining and Layer Freezing}
\label{sec:freeze}

To understand the impact of adaptive pretraining and layer freezing, we 
display zero-shot multilingual BERT results (test set) in 
\tabref{freeze_results}. We can see that there are clear benefits for 
adaptive pretraining (top-3 vs.\  bottom-3 rows). For layer freezing, we 
have 
3 options: no freezing (``$\emptyset$''), only freezing the subword 
  embeddings (``*'') and freezing the first 3 layers (``**'').  The 
second option (subword embedding frozen) consistently produces the best 
results (irrespective of whether adaptive pretraining is used), showing 
that this approach is effective in tackling the synchronisation issue 
(\secref{cross-lingual}) that arises when we fine-tune a multilingual 
model on one language.

\begin{table}[t]
\caption{\label{Tab:freeze_results} Influence of adaptive pretraining 
(``Ad. Pt.'') and layer freezing (``Frz.'') for results (Accuracy (\%)).  ``$\emptyset$'' denotes no 
freezing of any layers; ``*'' freezing the subword embedding layer; and  
``**'' freezing the first 3 layers.}
\begin{center}
\begin{adjustbox}{width=0.65\linewidth}
\begin{tabular}{cccccc}
\toprule
\textbf{Ad.} & \multirow{2}{*}{\textbf{Frz.}} & 
\textbf{T15/16}$\rightarrow$ & \textbf{PHEME}$\rightarrow$ & 
\textbf{WEIBO}$\rightarrow$ & \textbf{WEIBO}$\rightarrow$ \\
\textbf{Pt.} && \textbf{WEIBO} & \textbf{WEIBO} & \textbf{T15/T16} & 
\textbf{PHEME}\\
\midrule
\multirow{3}{*}{N} & $\emptyset$      &  $53.2$ & $58.4$ &$50.1$ & 
$54.5$ \\ &*      &  $61.3$ & $60.0$ & $52.6$ & $63.8$ \\ &**      & 
$57.5$ & $60.9$ & $52.9$  & $58.5$ \\ \hline
\multirow{3}{*}{Y} & $\emptyset$  & $56.6$ & $61.8$  & $50.9$ & $61.6$  
\\ &* & $\textbf{64.3}$ & $\textbf{65.9}$  & $\textbf{60.8}$ & 
$\textbf{67.2}$  \\ &** & $60.3$ & $63.8$ & $55.0$ & $61.8$  \\ 
\bottomrule
\end{tabular}
\end{adjustbox}
\end{center}
\end{table}

\begin{figure}[t]
\centering
  \includegraphics[width=0.75\columnwidth]{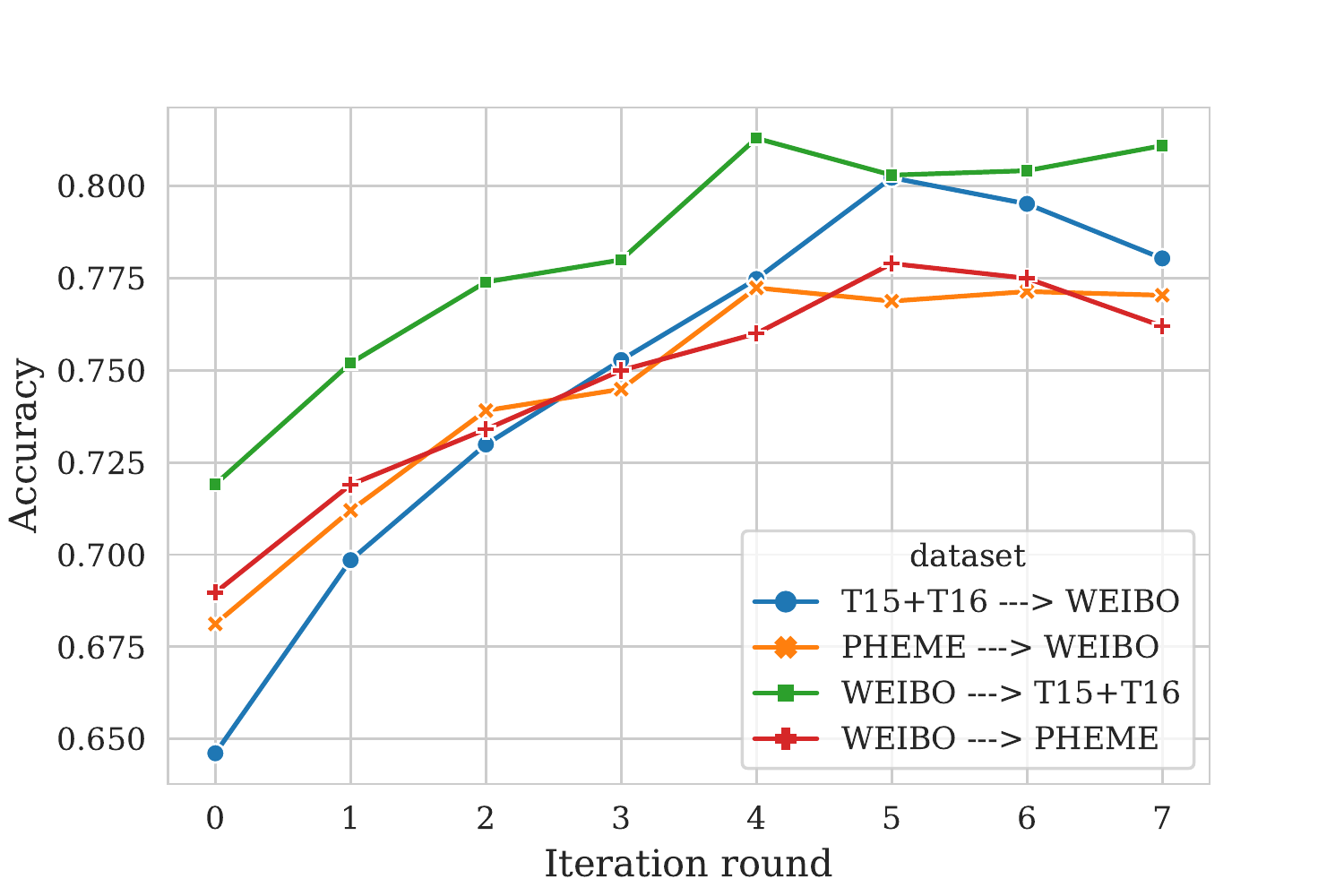}
  \caption{Accuracy over iteration during self-training.}
  \label{fig:xlmroberta_loop}
\end{figure}

\subsection{Self-training}

To measure the influence of the self-training loop, we present target 
performance (test set) of our multilingual model (``MF-XLMR$+$ST$+$GL'')
over different iterations in the self-training loop in 
\figref{xlmroberta_loop}. 
We can see the performance improves rapidly in the first few iterations, 
and gradually converges after 4--7 iterations. These results reveal the 
importance of refining the model over multiple iterations during 
cross-lingual transfer.

\begin{table}[t]
\caption{\label{Tab:few_shot_results} T15/16$\rightarrow$WEIBO results (Accuracy(\%)) 
as we incorporate more ground truth target labels (``GT Label'').}
\begin{center}
\begin{adjustbox}{width=0.42\linewidth}
\begin{tabular}{rcc}
\toprule
\% GT Label & Supervised & Zero-shot \\ \midrule
0\%      & --- & $80.2$       \\ 
20\%     & $79.8$ & $86.3$  \\ 
40\%     & $83.3$  & $89.2$ \\ 
60\%     & $89.3$  & $92.9$  \\ \
80\%     & $91.0$  & $93.5$   \\ 
100\%    & $92.2$ & --- \\
\bottomrule
\end{tabular}
\end{adjustbox}
\end{center}
\end{table}

\subsection{Semi-supervised Learning}
\label{sec:few-shot}

Here we explore feeding a proportion of ground truth labels in the 
target domain to our zero-shot model 
(``MF-XLMR$+$ST$+$GL'') and compare it to supervised multilingual 
model (``XLMR$+$both''). We present T15/16$\rightarrow$WEIBO 
results (test set) in \tabref{few_shot_results}. We can see that the gap 
shrinks by more than half (12.0 to 5.9 accuracy difference) with just 
20\% ground truth target label.
In general our unsupervised cross-lingual approach is also about 20\% 
more data efficient (e.g.\ supervised accuracy@40\% $\approx$ 
unsupervised accuracy@20\%).
Interestingly, with 60\% ground truth our model outperforms the fully 
supervised model.

\section{Discussion and Conclusions}
One criticism of the iterative self-training loop is that it suffers 
from poor initial prediction which could lead to a vicious cycle that 
further  degrades the student model.  The poor initial predictions 
concern appears to less of a problem in our task, as the pure zero-shot 
models (i.e.\ without self-training) appear to do reasonably well when 
transferred to a new language, indicating that the pretrained 
multilingual models (e.g.\ XLMR) are sufficiently robust. By further 
injecting the gold labels from the source domain during self-training, 
we hypothesise that it could also serve as a form of regularisation to 
prevent continuous degradation if the initial predictions were poor. Also, although our 
proposed transfer learning framework has only been applied to 
multilingual rumour detection, the architecture of the framework is 
general and applicable to other tasks.

To conclude, we proposed a zero-shot cross-lingual transfer learning 
framework to build a multilingual rumour detection model using only 
labels from one language. Our framework introduces: (1) a novel 
self-training loop that iteratively refines the multilingual model; and 
(2) ground truth labels in the source language during cross-lingual 
transfer. Our zero-shot  multilingual model produces strong rumour 
detection performance in both source and target language.

\section*{Acknowledgments}
This research is supported in part by the Australian Research Council Discovery Project DP200101441.

\end{document}